\documentclass[letterpaper, 10 pt, conference]{ieeeconf}  
\usepackage{amsmath}
\usepackage{graphicx}
\usepackage{caption}
\usepackage{subcaption}
\usepackage{comment}
\graphicspath{ {./figures/} }
\usepackage[font=small]{caption}

\IEEEoverridecommandlockouts                              

\title{\LARGE \bf
CobotAR: Interaction with Robots using Omnidirectionally Projected Image and DNN-based Gesture Recognition
}

\author{Elena Nazarova$^{1}$, Oleg Sautenkov$^{1}$, Miguel Altamirano Cabrera$^{1}$, Jonathan Tirado$^{1}$,  \\ Valerii Serpiva$^{1}$, Viktor Rakhmatulin$^{2}$, and Dzmitry Tsetserukou$^{1}$
\thanks{$^{1}$ The authors are with the Space Center and $^{2}$ Center for Design, Manufacturing and Materials, Skolkovo Institute of Science and Technology (Skoltech), 121205 Bolshoy Boulevard 30, bld. 1, Moscow, Russia. {\tt\small \{elena.nazarova, oleg.sautenkov, jonathan.tirado, miguel.altamirano, valerii.serpiva, viktor.rakhmatulin, d.tsetserukou\}@skoltech.ru\ }}
}

\begin{document}

\maketitle
\thispagestyle{empty}
\pagestyle{empty}

\begin{abstract}

Several technological solutions supported the creation of interfaces for Augmented Reality (AR) multi-user collaboration in the last years. However, these technologies require the use of wearable devices. We present CobotAR - a new AR technology to achieve the Human-Robot Interaction (HRI) by gesture recognition based on Deep Neural Network (DNN) - without an extra wearable device for the user. The system allows users to have a more intuitive experience with robotic applications using just their hands. The CobotAR system assumes the AR spatial display created by a mobile projector mounted on a 6 DoF robot. The proposed technology suggests a novel way of interaction with machines to achieve safe, intuitive, and immersive control mediated by a robotic projection system and DNN-based algorithm. We conducted the experiment with several parameters assessment during this research, which allows the users to define the positives and negatives of the new approach. The mental demand of CobotAR system is twice less than Wireless Gamepad and by 16\% less than Teach Pendant.

\end{abstract}

\section{Introduction}

Several technological solutions related to the creation of user interfaces for multi-user collaboration, education, maintenance, or safer and more effortless industrial operation (Microsoft Dynamics \cite{Microsoft}, OmniPack \cite{NativeRobotics}, RoMA \cite{10.1145/3173574.3174153}, Loki \cite{10.1145/3332165.3347872}) had emerged in the last years. The majority of them are based on Augmented Reality or Virtual Reality (AR-VR). Commonly, these technologies are complemented by additional devices that provide the visualization of virtual objects. They can be wearable devices, e.g., SixthSense \cite{10.1145/1667146.1667160}, AAR \cite{10.1145/3379337.3415849}, HoloLens \cite{Hololens}, Google Glass \cite{google}, or portable devices such as tablets or cellphones. These technologies provide good advantages and facilities that can contribute to the development of more Intuitive HRI for easy interaction between Robotic Systems and Humans. 

\begin{figure}[h!]
  \centering

  \includegraphics[width=1.0\linewidth]{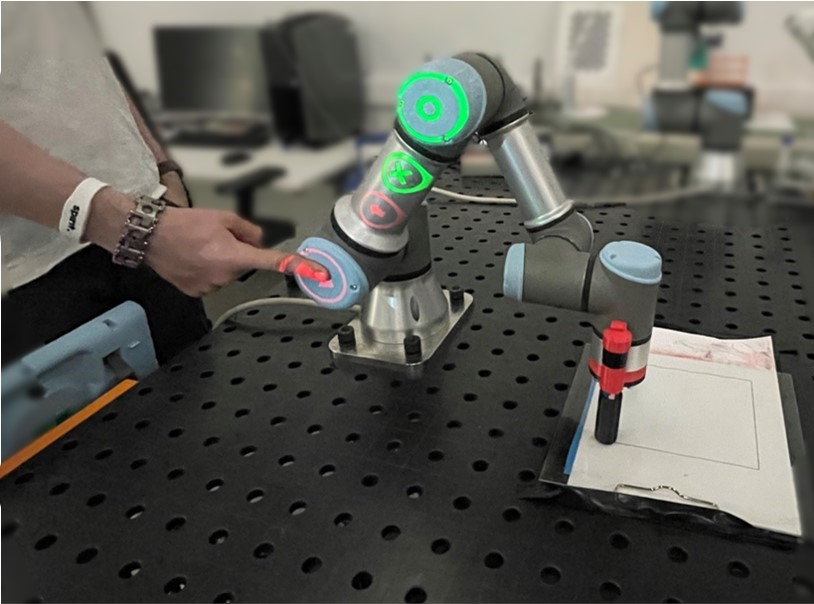}
  \caption{CobotAR interface by projection and DNN-based hand gesture recognition to control a collaborative robot UR3.}
  \label{fig:SystemUR3}
\end{figure}

Industry 4.0 trends show that Cyber-Physical Systems (CPS) are high-level linked with human interaction. That is one of the main reasons why the new technologies are devoted to support and emphasize this interaction \cite{KagermannWahlsterHelbig2013en}. HRI requires intuitive and efficient interfaces. User-Centered Design (UCD) systems decrease the complexity of human-Robot interfaces and increase the system's intuitive operation. The result is an apparent enhancement of CPS efficiency and a substantial reduction of human errors \cite{7819154}. AR technologies permit the creation of HRI interfaces in real environments, where the virtual and physical elements are incorporated together \cite{OSTANIN2019695}. AR can reduce the gap between simulation and implementation by enabling the prototyping of algorithms on a combination of physical and virtual objects where robots and humans can interact in the same environment \cite{7354138}.

Projection mapping-related technologies are used to augment the interface with projected elements. Thanks to projection mapping, we can generate visual illusions by changing the shapes of real objects or adding extra features. However, when projection mapping is applied in dynamic systems, it has critical limitations in the process latency and the target tracking. Additionally, if the capture-projection system is installed in a static position, the projection area is also restricted \cite{10.1145/3415255.3422896}.

The interactive projection technology introduced by Mistry et al. \cite{10.1145/1667146.1667160} allows users to interact with projected objects or images almost on any surface.  The hand movements and gestures are tracked and recognized by the camera and hand-worn marker. This technology allows users to interact with the projection on different surfaces, such as navigating a map, drawing on any surface, and displaying and showing photos. However, the interaction is only with the projected information and is required to use a color marker on the user's finger to track the hand gestures.

Hartmann et al. \cite{10.1145/3379337.3415849} introduced a similar technology by combining a head-mounted actuated pico projector with a Hololens AR headset. Demonstrations showcase ways to use head-mounted displays and the projected together, such as distributing content across depth surfaces, expanding the field of view, and enabling bystander collaboration.

Projector-based AR is one of the most suitable approaches for HRI applications because of two main advantages. Firstly, the user's environment is augmented with images that are integrated directly into the real objects, not only in their visual field \cite{10.1145/1185657.1185796}. Secondly, this technology allows the user to have more natural interaction with the robot since the person is not wearing or using any tracking or projection device on his body.

We introduce CobotAR, a novel HRI system that implements AR projector-based spatial displays by a Camera-Projector Module (CPM) mounted on a 6 DoF collaborative robot, and DNN-based gesture recognition to interact with a collaborative robot. In Section II, an overview of the system is presented, where the system's main components are described. An experimental evaluation is presented in Section III, where the proposed system was compared with other interfaces to control a collaborative robot and evaluate its advantages and limitations.     

\section{System Overview}
The CobotAR is a novel HRI system containing AR projector-based spatial displays, which generates an interactive projected Graphic User Interface (GUI) on a UR3 collaborative robot. Using DNN-based gesture recognition, the system tracks the position of the user’s hand, it allows users to interact with the projected GUI and control the position of the robot in an intuitive way. The projector-camera module is mounted on a second robot, not fixed statically, to provide a bigger projection area and avoid the shadows in the projection.

\begin{figure}[h!]
  \centering
  \vspace{0.2 cm}
  \includegraphics[width=1.0\linewidth]{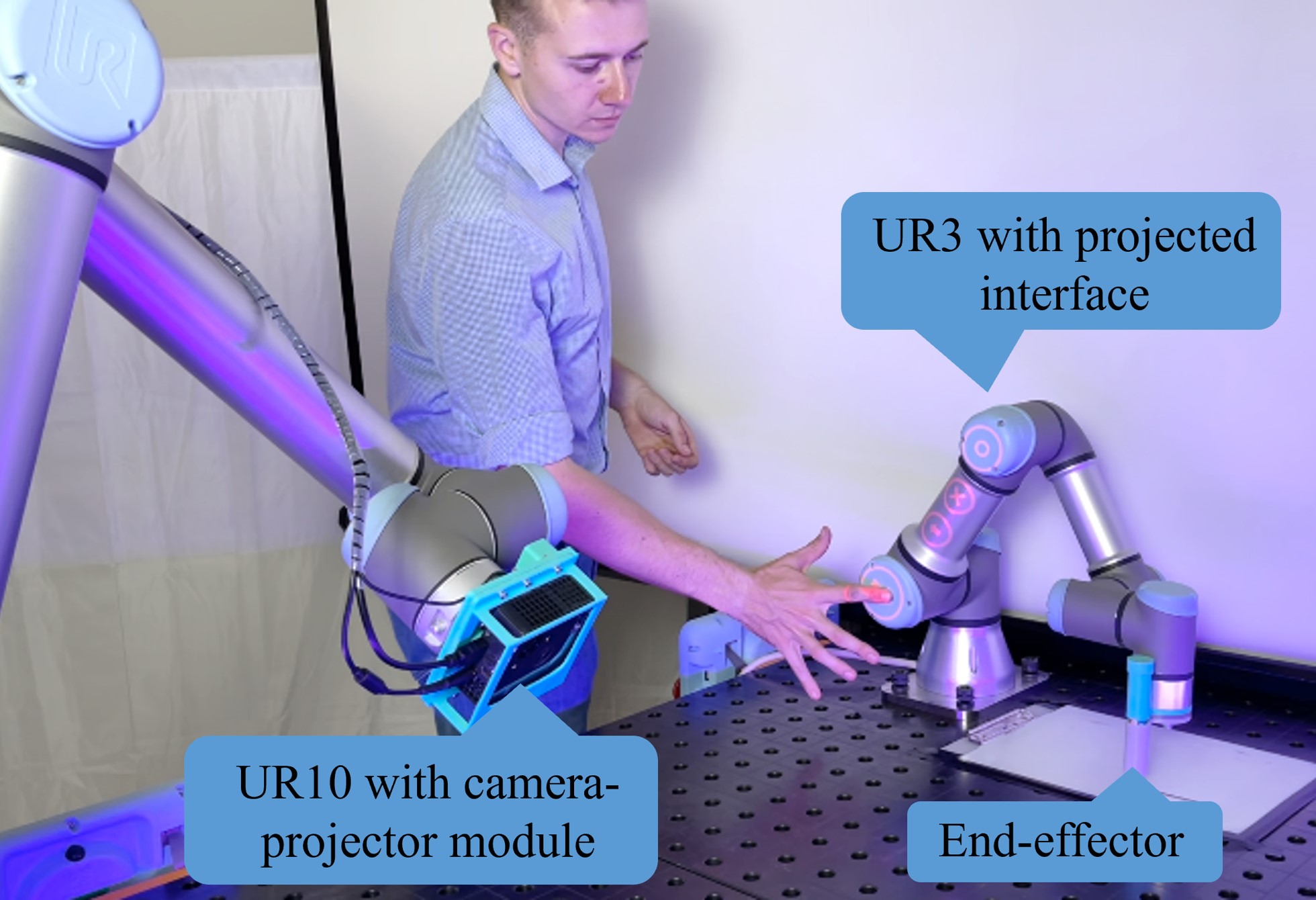}
  \caption{User controls the UR3 collaborative robot through projected interface from ultra-compact mobile projector Dell M115HD and Logitech HD Webcam C930e mounted on the end-effector of a robot UR10.}
  \label{fig:System architecture}
\end{figure}

\begin{figure}[h!]
  \centering
  \vspace{0.3 cm}
  \includegraphics[width=1\linewidth]{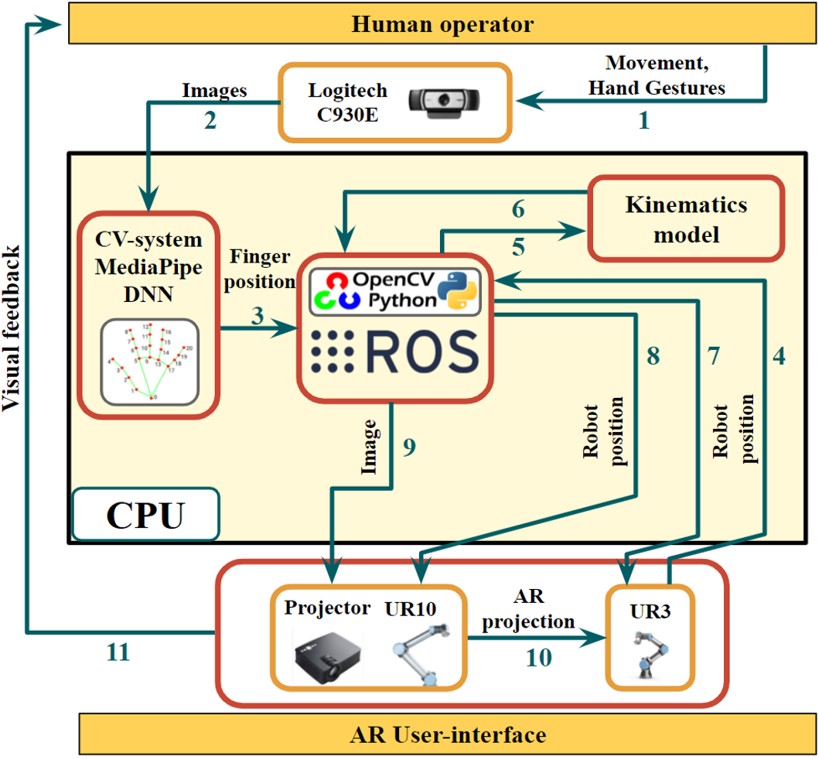}
  \caption{System architecture.}
  \label{fig:System overview}
\end{figure}

The hardware components of the CobotAR system are a mobile projector Dell M115HD, a Logitech HD Webcam C930e, two 6 DoF collaborative robots from Universal Robots UR3 and UR10, and a desktop computer. Three computational modules are responsible for data processing and robot controller: a) gesture recognition based on DNN, b) image processing through OpenCV library, and c) URX python library to communicate with Universal Robots. The system architecture is shown in Fig. \ref{fig:System architecture}, and system overview is shown in Fig. \ref{fig:System overview}. 

The DNN-based algorithm detects the user's hand gestures while the CPM follows the target UR3 robot position. It allows the users to interact with any type of GUI projection and to control the robot's position more intuitively. The CPM is mounted on a UR10 robot to provide a more extensive projection area and avoid the shadows in the projection. It follows constantly the UR3 robot target surface position to maintain the right projection. Simultaneously, the Computer Vision (CV) algorithm processes the Webcam image using the hand tracking module. It estimates the position of the fingers in a specific area and defines it as a "press button" command. As a result of the gesture recognition process, the central program defines the required robotic action. CobotAR can present different GUI with buttons and interactive robotic parts (user visualize the inner structure of the robot without disassembling it).

\subsection{Robots kinematics}

The GUI is projected on the UR3 robot, as shown in Fig. \ref{fig:CobotAR system controlling}. The UR10 robot, with the CPM, continuously follows the position and orientation of the middle point of the projected link, calculated by inverse kinematics. The position and orientation of the UR10 are calculated by forward kinematics. The image of the GUI does not change with the motion. 

The Denavit-Hartenberg (DH) parameters of the UR3 robot are represented in Table \ref{table1},
where Joint N is each joint in the robot, where $\Theta$ is the joint variable, $a$, $d$ and $\alpha$ are the fixed link parameters. \cite{Craig}

Table \ref{table2} describes DH parameters of our system for the GUI projection. 

We process received values from UR3 robot joints and get the final transformation matrix. We extract the position and orientation of the desired point\cite{Craig}. After all, we transfer it to the UR10 robot with projector-camera system.

\begin{table}[h]
\vspace{0.5cm}
\caption{Denavit-Hartenberg representation of UR3}
\label{table1}
\begin{center}
\begin{tabular}{|c||c||c||c||c|}
\hline
Joint & $\Theta$ & a & d & $\alpha$\\
\hline
Joint 1 & $\Theta_{1}$ & 0 & 0.15185 & $\pi/2$ \\
\hline
Joint 2 & $\Theta_{2}$ & -0.24355 & 0 & 0\\
\hline
Joint 3 & $\Theta_{3}$ & -0.2132 & 0 & 0\\
\hline
Joint 4 & $\Theta_{4}$ & 0 & 0.13105 & $\pi/2$\\
\hline
Joint 5 & $\Theta_{5}$ & 0 & 0.08535 & $-\pi/2$\\
\hline
Joint 6 & $\Theta_{6}$ & 0 & 0.0921 & 0\\
\hline

\end{tabular}
\end{center}
\end{table}

\begin{table}[h]
\vspace{0.5cm}
\caption{Denavit-Hartenberg representation of the model}
\label{table2}
\begin{center}
\begin{tabular}{|c||c||c||c||c|}
\hline
Joint & $\Theta_{n}$ & a & d & $\alpha$\\
\hline
Joint 1 & $\Theta_{1}$ & 0 & 0.15185 & $\pi/2$\\
\hline
Joint 2 & $\Theta_{2}$ & -0.12176 & 0.40 & 0\\
\hline

\end{tabular}
\end{center}
\end{table}

The UR10 is initially positioned to project GUI on the UR3 link. Further motions were conducted relative to this reference position.

\subsection{DNN-based gesture recognition}

The DNN-based gesture recognition module is implemented on the basis of the Mediapipe framework. It provides high-fidelity tracking of the hand by employing Machine Learning (ML) to infer 21 key points of a human hand per a single captured frame.

If the index finger coordinates are located in the button's area, and the gesture has been changed from "Palm" to "One", the corresponding button will activate. The algorithm sends a number of activated buttons by ROS framework to the robot control system.

DNN was trained to recognize 8 gestures, two gestures were chosen to perform pressing the buttons.

\section{Experimental Evaluation}
The principal approach of this paper is to design a new AR-technology to control a collaborative robot by a projected interface and DNN-based gesture recognition without an extra wearable device for the user.
We conducted a series of user evaluations to determine the advantages and disadvantages of the proposed system in comparison with existing interfaces. The interfaces to compare were the Teach Pendant from UR3 and the Logitech F710 Wireless Gamepad \cite{Gamepad}.

\subsection{Experimental Design}

Our team designed, printed using a 3D printer, and assembled a marker holder on the end-effector of the UR3 robot before the experiment. Moreover, we fixed a 150x150 mm square printed on white paper on the table where the robots are located. Participants were asked to repeat the printed path of the 150x150 mm square trajectory manipulating by the Teach Pendant, the Logitech F710 Wireless Gamepad, and the novel CobotAR system to evaluate the controllability of the robot by the three different interfaces.

We used data such as the coordinates of each participant's TCP UR3 robot trajectory and the comparison time for the experimental evaluation. Moreover, each participant completed a questionnaire for each interface based on The NASA Task Load Index (NASA-TLX) in order to determine the more user-friendly system and give some comments about the advantages and disadvantages of the use of CobotAR system relative to the other interfaces. 

Twelve participants (3 females) volunteering conducted the test, aged from 22 to 43 years. None of them reported any problems and difficulties in CobotAR functionality.

Before starting the experiment, the participants performed a training session, where each participant familiarized himself/herself with each interface and tested it. After the training session, the UR3 robot returned to the predetermined initial position, and the experiment started.

\subsection{CobotAR system controlling}
\begin{figure}[ht!]
  \vspace{0.3cm}
  \includegraphics[width=1.0\linewidth]{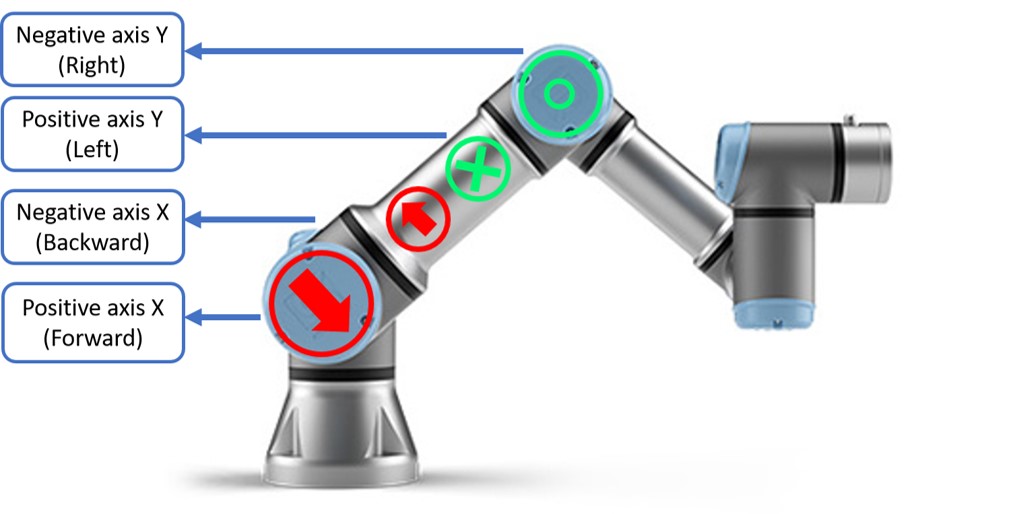}
  \caption{CobotAR control interface. The interface contains four buttons. Each button controls one direction of the robot in the XY plane. The gesture recognition system detects the pressed button and moves the robot in the desired direction.}
  \label{fig:CobotAR system controlling}
\end{figure}

The GUI to control the UR3 robot through CobotAR system is represented in Fig. \ref{fig:CobotAR system controlling}. By pressing on the projected green buttons, the robot TCP moves in the y-axis, with a positive or negative direction, respectively. The circle in the top simulates the front part of an arrow, and the cross the back part of an arrow, to specify the direction of the TCP movement. By pressing on the projected red buttons, the robot TCP moves in the x-axis, with a positive or negative direction according to the direction of the arrows. The user can use only one button simultaneously. During the experiment, the user can correct each time the trajectory of the robot. 

\subsection{UR3 Teach Pendant}
\begin{figure}[h!]
  \vspace{0.5cm}
  \includegraphics[width=1.0\linewidth]{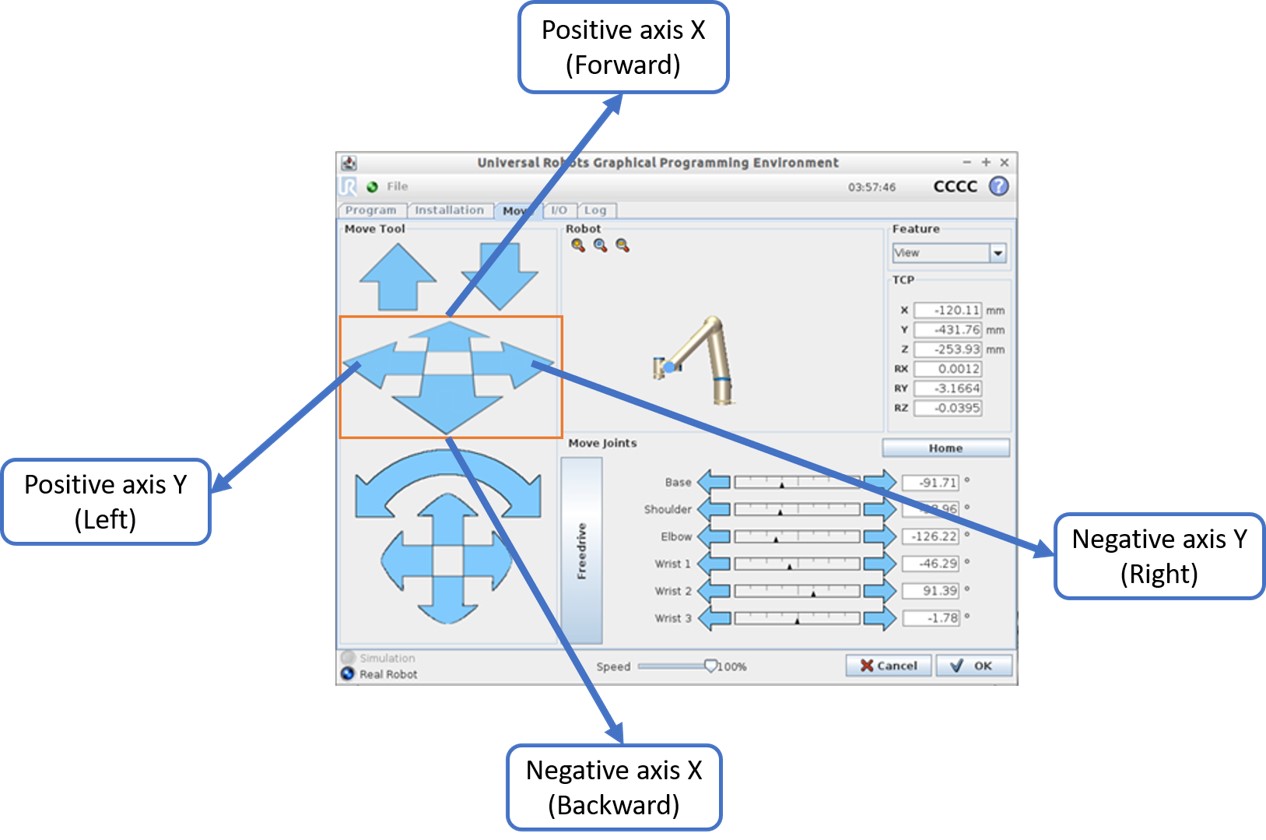}
  \caption{Teach Pendant control interface. It is the original UR control interface. The touch Teach Pendant contains several buttons and functions through which the user can manipulate each of the 6-DoF of the robot.}
  \label{fig:Robot Teach Pendant system controlling}
\end{figure}

The interface to control the UR3 robot TCP  through the Teach Pendant is represented in Fig. \ref{fig:Robot Teach Pendant system controlling}. By pressing on the buttons in the orange box, user has the ability to move the robot TCP in positive and negative X and Y directions. The user can use only one button simultaneously. During the experiment, the user could correct the trajectory of the robot. 

\subsection{Logitech F710 Wireless Gamepad}
\begin{figure}[h!]
  \vspace{0.5cm}
  \includegraphics[width=1.0\linewidth]{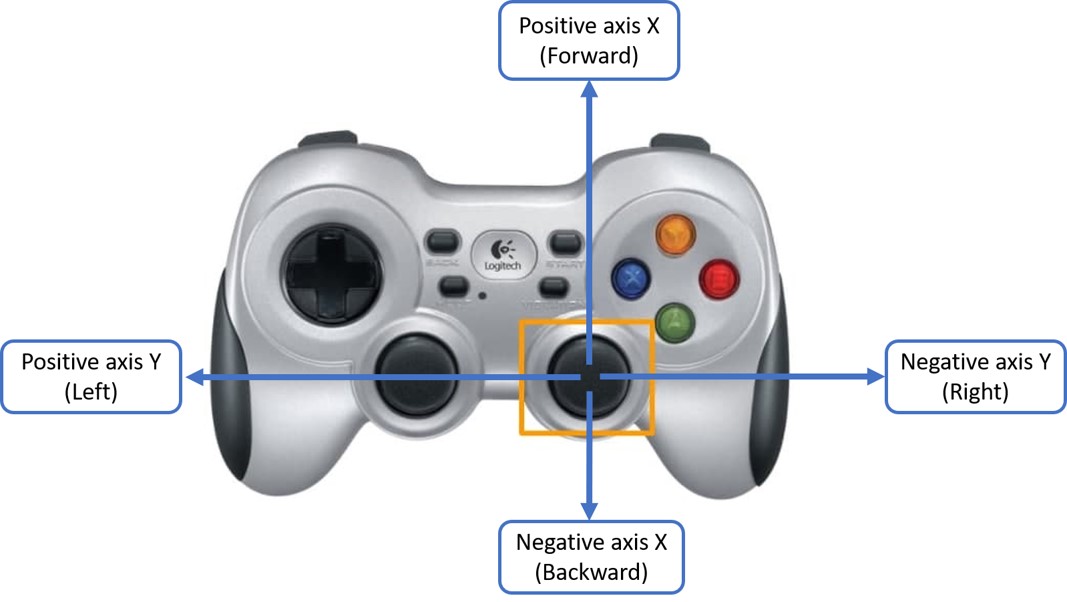}
  \caption{Wireless Gamepad interface. The interface uses the thumb Wireless Gamepad of a Logitech Wireless Gamepad F710. This device allows the user to control the movements of the UR robot in the plane XY.} 
  \label{fig:Wireless Gamepad system controlling}
\end{figure}

The Wireless Gamepad stick was used to control the position of the robot TCP as shown in Fig. \ref{fig:Wireless Gamepad system controlling}. The position of the TCP changes in the x and y axis by the movement of the right stick in the orange box. The system works in the Wireless Gamepad control way. The user could change robot movement not only in a straight lines, but also in diagonal.

\section{Experimental Results}

\begin{figure}[h!]

  \includegraphics[width=1.0\linewidth]{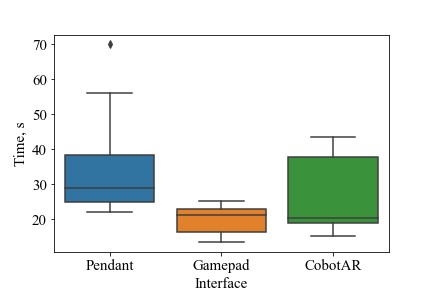}
  \caption{The time performance comparison of the participants work on each interface.}
  \label{fig:TimeComparison}
\end{figure}

Figure \ref{fig:TimeComparison} presents a summary of the participants’ time spent to draw the square using each interface by boxplots. Considering average values, the fastest interface is the Logitech F710 Wireless Gamepad (20,07s), second the CobotAR system (27,61s), and third the UR3 Teach Pendant (38,16s). 

\begin{figure}[h!]

  \includegraphics[width=1.0\linewidth]{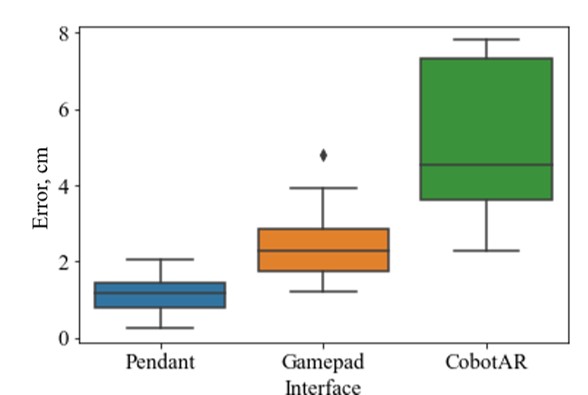}
  \caption{The average error between the printed square and the trajectory designed by the users using each of the interfaces.}
  \label{fig:ErrorComparison}
\end{figure}

In order to evaluate the statistical significance of the differences between the time from each of the interfaces, we analyzed the results using single factor repeated-measures ANOVA, with a chosen significance level of $\alpha<0.05$. According to the ANOVA results, there is a statistical significant difference in the operation time in each of the interfaces, $F(2,36) = 6.9962, p = 6.03\cdot10^{-3}$. The ANOVA showed that the interface used to control the robot significantly influence the time of the operation.

The paired t-tests showed statistically significant differences between the time from the use of the proposed system CobotAR and the Logitech F710 Wireless Gamepad ($p=0.0468 < 0.05$), between the time from the Logitech F710 Wireless Gamepad and the UR3 Teach Pendant ($p=2.9\cdot10^{-3} < 0.05$). 

We see that the UR3 Teach Pendant presents the longest time because of the interaction between the user's finger and the resistive sensor display, which has more delay and miss-clicks than the other interfaces. The Logitech F710 Wireless Gamepad is the fastest because the user does not need to spend the time to switch the direction with an analog stick sensor. CobotAR stands between them.

The average error between the printed square and the trajectory drawed by the users using each of the interfaces is presented in Fig. \ref{fig:ErrorComparison}. Participants were, on average, 1.1 cm from the right path using the UR3 Teach Pendant interface. Using the Logitech F710 Wireless Gamepad, the average error was 2.5 cm, and using CobotAR, it was 5.1 cm. 

\begin{figure}

\begin{subfigure}[h]{0.5\linewidth}
\includegraphics[width=\linewidth]{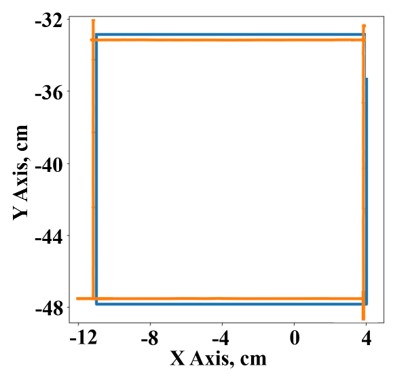}
\caption{CobotAR graph.}
\end{subfigure}
\hfill
\begin{subfigure}[h]{0.49\linewidth}
\includegraphics[width=\linewidth]{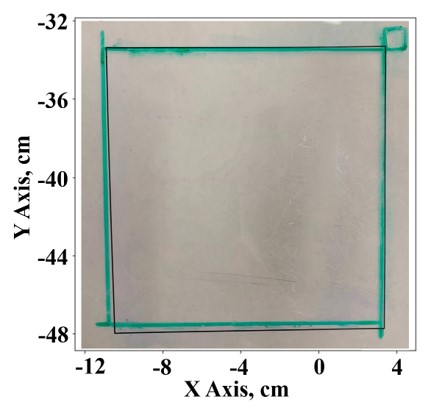}
\caption{CobotAR drawing.}
\end{subfigure}%
\caption{End effector trajectory followed by CobotAR interface.}
\label{fig:CobotAR results}

\vspace{0.5cm}
\begin{subfigure}[h]{0.5\linewidth}
\includegraphics[width=\linewidth]{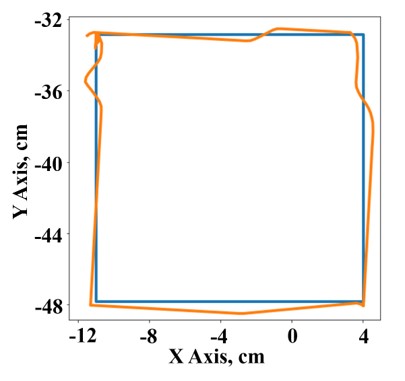}
\caption{Logitech F710 Wireless Gamepad graph.}
\end{subfigure}
\hfill
\begin{subfigure}[h]{0.49\linewidth}
\includegraphics[width=\linewidth]{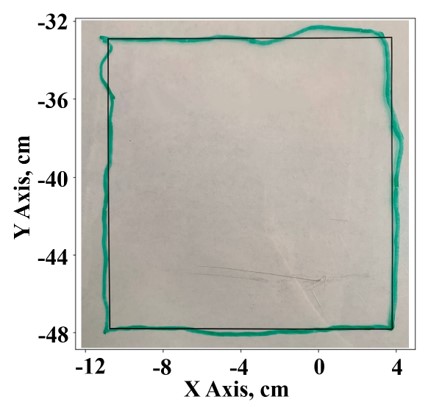}
\caption{Logitech F710 Wireless Gamepad drawing.}
\end{subfigure}%
\caption{End effector trajectory followed by Logitech F710 Wireless Gamepad.}
\label{fig:Wireless Gamepad results}

\vspace{0.5cm}
\begin{subfigure}[h]{0.5\linewidth}
\includegraphics[width=\linewidth]{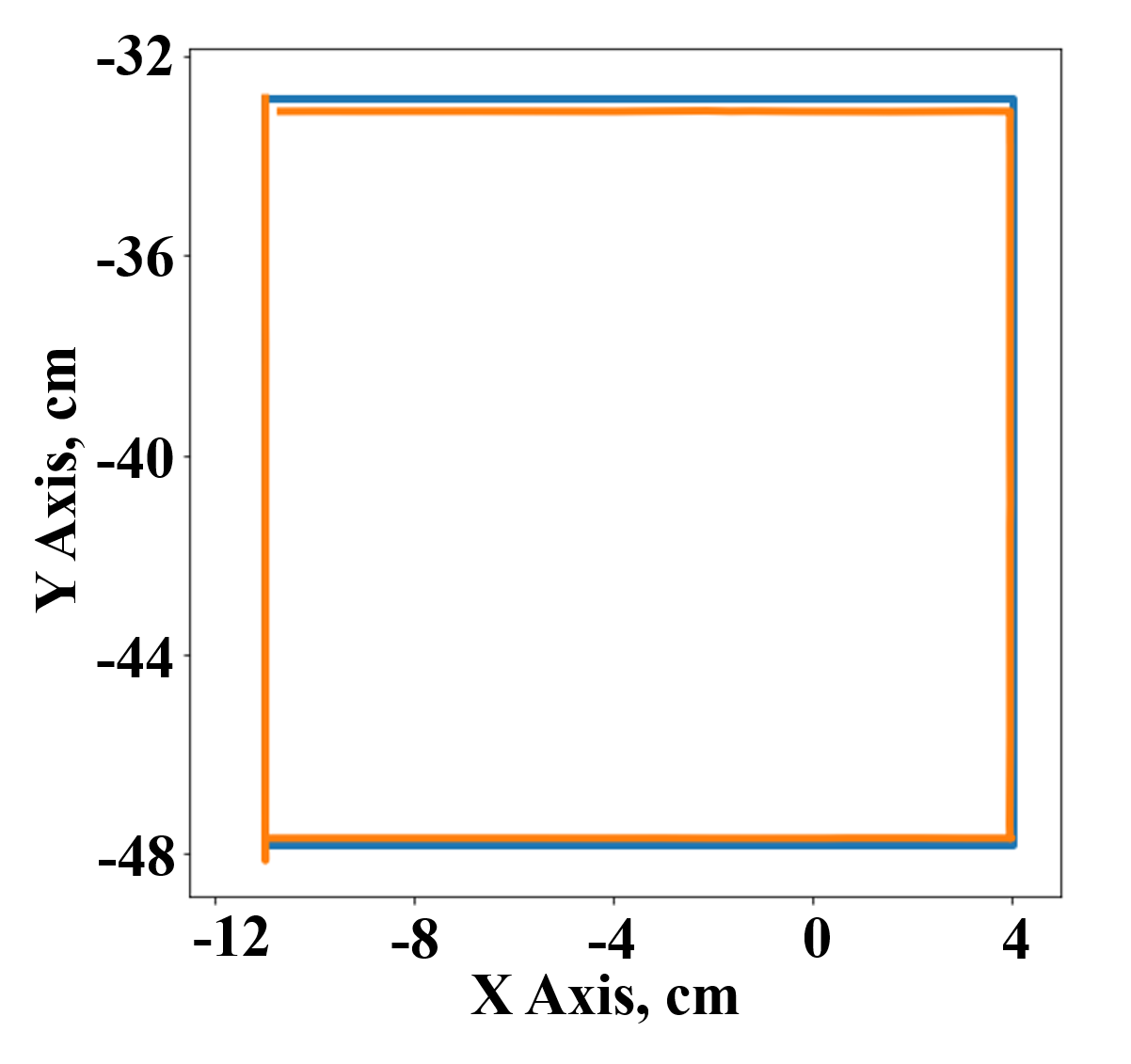}
\caption{UR3 Teach Pendant graph.}
\end{subfigure}
\hfill
\begin{subfigure}[h]{0.49\linewidth}
\includegraphics[width=\linewidth]{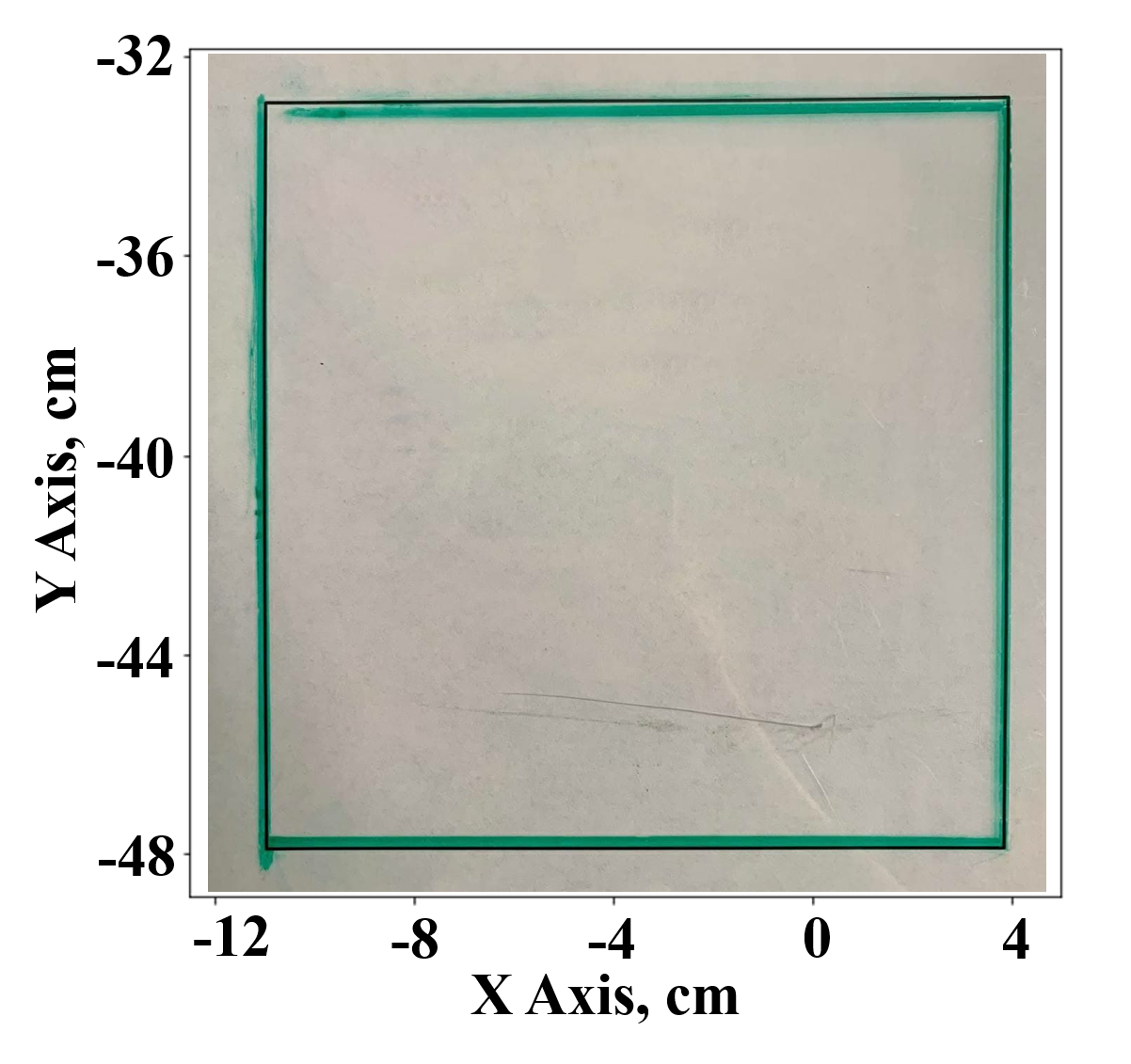}
\caption{UR3 Teach Pendant drawing.}
\end{subfigure}%
\caption{End effector trajectory followed by UR3 Teach Pendant interface.}

\label{fig:Teach pendant results}
\end{figure}

Fig. \ref{fig:CobotAR results}, Fig.\ref{fig:Wireless Gamepad results}, and Fig.\ref{fig:Teach pendant results} show the graphs of the ideal trajectory of a square and the real trajectory of user 1, which he drew using the different interfaces. The data shown in these graphs were captured during the experiment of user 1.

\begin{table}[h]
\caption{NASA TLX rating for each interface}
\label{table_NASA}

\begin{center}
\scalebox{0.9}{%
\begin{tabular}{|c||c||c||c|}
\hline
  & Teach pendant & Wireless Gamepad & CobotAR\\
\hline
Mental Demand & 3,92 & 6,08 & 2,83\\
\hline
Physical Demand & 4,25 & 5,33 & 3,08\\
\hline
Temporal Demand & 4,25 & 4,58 & 4,08\\
\hline
Overall performance & 2,00 & 5,58 & 2,17\\
\hline
Effort & 4,17 & 5,33 & 3,50\\
\hline
Frustration & 4,00 & 5,92 & 3,67\\
\hline
Average time & 34,75 & 20,08 & 27,34\\
\hline
STD(mm) & 1,1 & 2,5 & 5,1\\
\hline
Average TLX & 32,26 & 46,90 & 27,62\\
\hline

\end{tabular}}
\end{center}
\end{table}

\subsection{Average score of NASA TLX rating}

Table \ref{table_NASA}:NASA TLX rating showed the results of the NASA TLX rating for each interface. Results showed that participants had the lowest mental workload using the CobotAR system (27,6 out of 100), whereas the UR3 Teach Pendant interface mental workload (32 out of 100) was slightly higher. The Logitech F710 Wireless Gamepad had the worst conditions (47 out of 100) according to the NASA TLX rating.

\subsection{Post-Experience Questionnaire}

Generally, people were very inspired by CobotAR system. Some of the participants noticed that they could push the robot only by finger in a desirable direction. 
The comment: "I liked the projector interface (CobotAR), it felt like I am pushing the robotic arm, and it is moving. If it would be a bit faster, it will feel even better."

Some of the participants were confused about dot and cross on the buttons and suggested to use different signs:
"The dot and cross may not be very intuitive; you can replace them with the words: "on" and "from" or something else."

"I enjoyed driving the CobotAR. It was both engaging and exciting. I want to try to manage it a couple more times. It involves you directly in the operation."

 The main advantage of this interface is that a person can move the robot with one hand, and the second hand is free. Secondly, it is non-obligatory to switch constantly the view from interface to end-effector. The person can observe them simultaneously. Third, the operator gets a kind of haptic feedback from the robot, and he can feel where he is moving the robot's hand.

\section{Applications}

\begin{figure}
\vspace{0.2cm}
\begin{subfigure}[h]{0.475\linewidth}

\includegraphics[width=\linewidth]{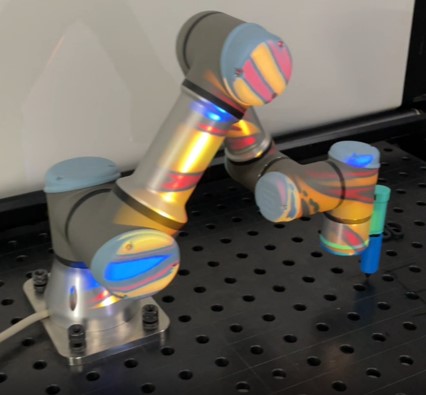}
\caption{CobotAR.Skin}
\end{subfigure}
\hfill
\begin{subfigure}[h]{0.495\linewidth}
\includegraphics[width=\linewidth]{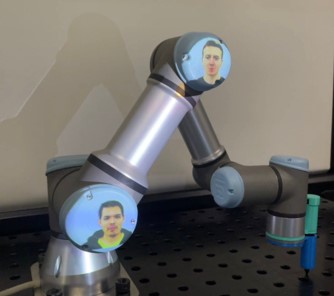}
\caption{CobotAR.Zoom}
\end{subfigure}%

\vspace{0.6cm}
\begin{subfigure}[h]{0.49\linewidth}
\includegraphics[width=0.97\linewidth]{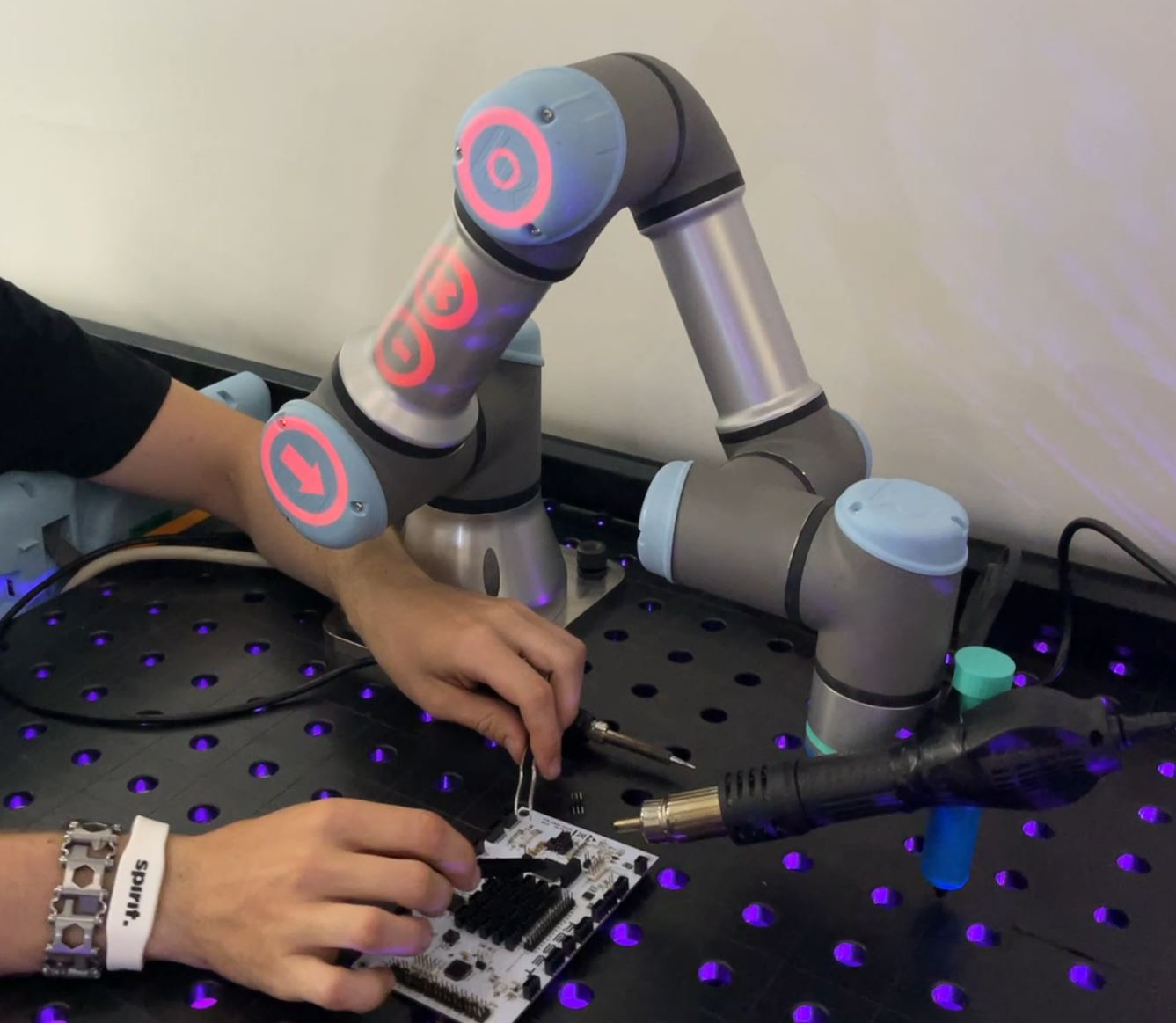}
\caption{CobotAR.Soldering}
\end{subfigure}
\hfill
\begin{subfigure}[h]{0.49\linewidth}
\includegraphics[width=\linewidth]{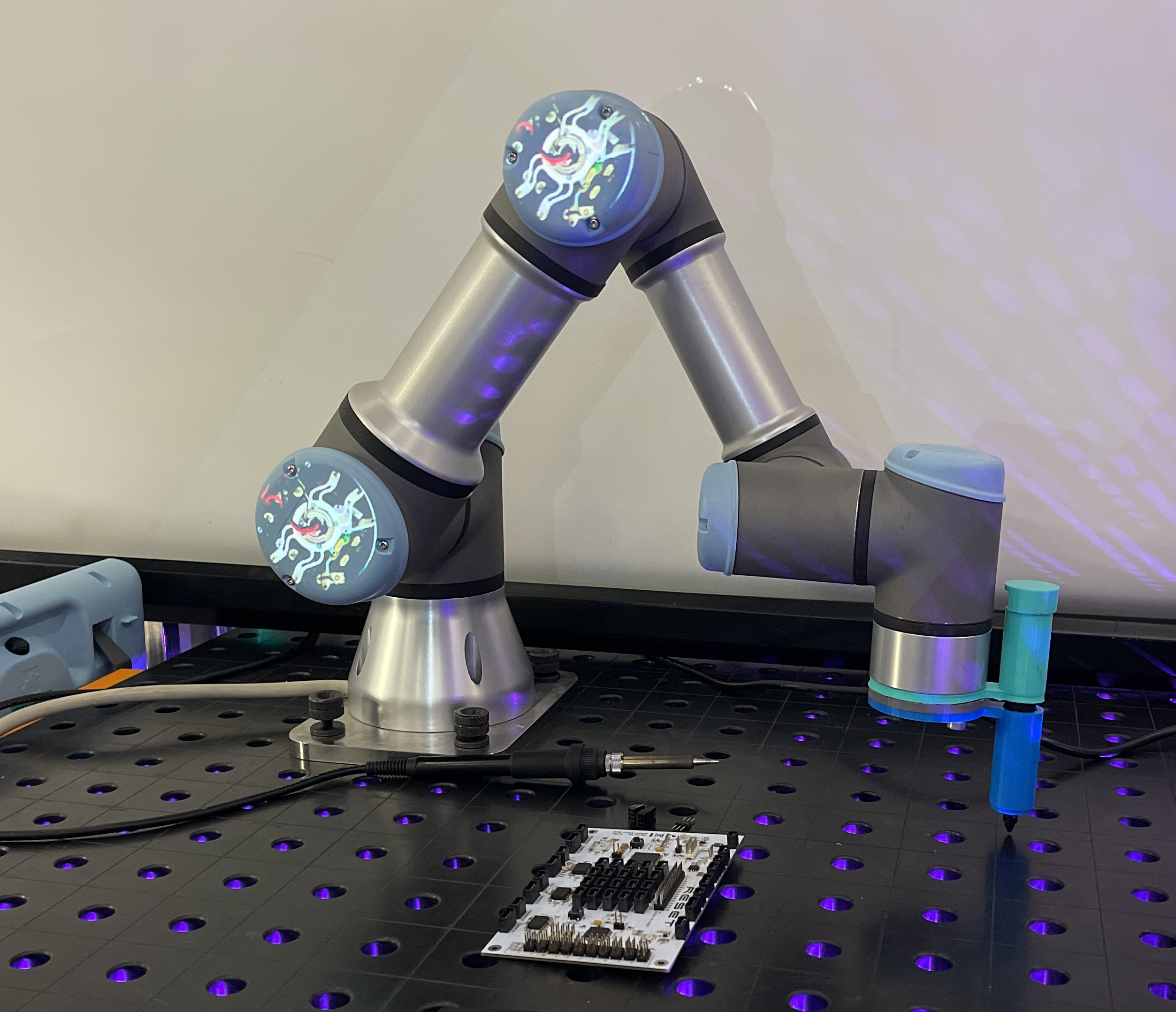}
\caption{Cobotar.Education}
\end{subfigure}%
\vspace{0.3cm}
\caption{Applications of CobotAR technology. }
\label{fig:Applications}

\end{figure}

The CobotAR technology could be potentially used in various applications: interactive education programs and entertainment, industrial human-robot collaborative scenarios as shown in Figure \ref{fig:Applications}. 
   
a) CobotAR.Skin is a dynamically projected skin on the robot's surface, reflecting the robot's state and its current task. For instance, the skin may indicate the proximity to the robot in collaborative tasks.

b) CobotAR.Zoom is a robot repairing support service based on online teleconferencing. The faces of the supporting team and relevant information are projected on the robot joints.

c) CobotAR.Soldering is an AR interface to support collaborative soldering. The robot equipped with a fan could be easily moved in the workspace and the human worker could perform operations with two hands. 
d) CobotAR.Education is an AR system for immersive robotic lectures. The application allows a teacher to visualize the interactive internal parts of the robot using hand gestures.

Besides this, CobotAR could provide engaging activities on robotic exhibitions and demonstrations. For instance, CobotAR.Drawing could be a simple interface, where a user guides the robot with hand gestures to draw a name or any shape on the whiteboard by a marker.

\section{Conclusion and Future Work}
This paper introduced a novel HRI approach based on the graphical interface, projected directly on the robot's links, and DNN-based gesture recognition. The supporting robot move and orients the projected interface according to the movements of the controlled robot. The developed approach does not require an application of any wearable devices. We performed a controlled experiment among three interfaces to compare the overall performance, ease of use, and acceptance rating.

The CobotAR interface provides immersive and intuitive control to the user. Using CobotAR, participants achieved the best results at Nasa TLX ratings, such as Mental Demand, Physical Demand, Temporal Demand, Effort, and Frustration. Besides, the approach demonstrated accuracy and time performance comparable to other interfaces. This suggests that the CobotAR is suitable for robot operation, especially for demonstration and education. The mental demand of the CobotAR system is twice less than Wireless Gamepad and by 16 percent less than Teach Pendant.

For future work, we consider improving the gesture recognition algorithm and decreasing system delay to improve the CobotAR time performance and accuracy. Moreover, we plan to develop a virtual scene with two digital twins in Unity Engine, allowing us to both simulate and adjust the behavior of the robotic arms in HRI scenarios as considered in \cite{10.1145/3359996.3365049}.

\section*{Acknowledgment}
The reported study was funded by RFBR and CNRS according to the research project No. 21-58-15006.

\addtolength{\textheight}{-12cm}   








\end{document}